\documentclass[11pt]{article}
\usepackage{times}
\usepackage[margin=1in]{geometry}
\usepackage{graphicx} 
\usepackage[numbers,sort&compress]{natbib}
\usepackage{parskip}
\usepackage{xcolor}
\usepackage{hyperref}
\usepackage{authblk}

\newcommand\blfootnote[1]{%
  \begingroup
  \renewcommand\thefootnote{}\footnotetext{#1}%
  \endgroup
}

\title{\textbf{{\Large Operationalizing the Blueprint for an AI Bill of Rights: \\Recommendations for Practitioners, Researchers, and Policy Makers}}\\ \textit{\large{Working Paper\footnote{This is a work-in-progress. Please direct any feedback  to \href{mailto:aoesterling@g.harvard.edu}{aoesterling@g.harvard.edu} and \href{mailto:usha_bhalla@g.harvard.edu}{usha\_bhalla@g.harvard.edu}}}}}
\date{}
\author[1]{Alex Oesterling$^\mathsection$}
\author[1]{Usha Bhalla$^\mathsection$}
\author[2]{Suresh Venkatasubramanian}
\author[3]{Himabindu Lakkaraju}

\affil[1]{John A. Paulson School of Engineering and Applied Science, Harvard University}
\affil[3]{Harvard Business School, Harvard University}
\affil[2]{Center for Tech Responsibility, Brown University}

\begin{document}

\maketitle

\blfootnote{$\mathsection$: Equal contribution, order by coin flip.}

\begin{abstract}
As Artificial Intelligence (AI) tools become ubiquitous and are increasingly employed in diverse real-world applications, there has been significant interest in regulating these tools. To this end, several regulatory frameworks have been introduced by different countries worldwide. For example, the European Union recently passed the AI Act, and the White House issued an Executive Order on safe, secure, and trustworthy AI, associated guidance from the Office of Management and Budget, all of which built on the Blueprint for an AI Bill of Rights issued by the White House Office of Science and Technology Policy. Many of these frameworks emphasize the need for auditing and improving the trustworthiness of AI models and tools, underscoring the importance of safety, privacy, explainability, fairness, and human fallback options. Although these regulatory frameworks highlight the necessity of enforcing the aforementioned principles, practitioners often lack detailed guidance on implementing them. While there is extensive research on operationalizing each of these aspects, it is frequently buried in technical papers that are difficult for practitioners to parse. In this write-up, we address this shortcoming by providing an accessible overview of the various approaches available in existing literature to operationalize regulatory principles. For the sake of clarity and precision, we will specifically focus on the Blueprint for an AI Bill of Rights throughout this document. We provide easy-to-understand summaries of state-of-the-art literature and highlight various gaps that exist between the regulatory guidelines in the Blueprint for an AI Bill of Rights and existing AI research, including the trade-offs that emerge when operationalizing each of the highlighted principles. We hope that this work not only serves as a starting point for practitioners interested in learning more about operationalizing the regulatory guidelines outlined in the Blueprint for an AI Bill of Rights but also provides researchers with a list of critical open problems whose resolution can help bridge the gaps between regulations and state-of-the-art AI research. Finally, we note that this is a working paper and we invite feedback in line with the purpose of this document as described in the introduction.

\end{abstract}

\section{Introduction}
Over the past decade, AI tools have become ubiquitous, finding applications in various real-world settings, including healthcare, finance, education, and e-commerce. For instance, in healthcare, AI assists in diagnostics and personalized treatment plans, while in finance, it enhances fraud detection and risk management. The emergence of generative AI in recent years has further accelerated the adoption of AI technologies across multiple domains, leading to innovations such as automated content creation, advanced language models, and personalized learning systems. However, the widespread use of AI tools has also brought about significant challenges and risks. Concerns about privacy, biases and lack of transparency in AI decision-making, and the security of AI systems have become increasingly prominent. For example, AI systems can inadvertently perpetuate biases present in training data, leading to unfair treatment in hiring processes or loan approvals. Moreover, the potential misuse of AI for malicious purposes, such as deepfakes or automated cyber-attacks, poses substantial security threats.

To address these concerns, several regulatory frameworks have been introduced. Most notable among them are the European Union's (EU) AI Act \citep{euaiact}, the White House's Executive Order on safe, secure, and trustworthy AI (and associated guidance from the Office of Management and Budget) \citep{biden2023executive}, which builds on the Blueprint for an AI Bill of Rights \citep{aibillofrights} put forth by the United States White House Office of Science and Technology Policy (OSTP) as well as the AI Risk Management Framework \citep{airmf} developed by the National Institute for Standards and Technology. These are pivotal frameworks aimed at ensuring the ethical and responsible use of technology. 
The EU AI Act seeks to promote the deployment of trustworthy and safe AI systems while ensuring the protection of human rights against potential harms of AI. It provides guidelines for responsible development as well as placement of AI systems on the market and in services \citep{euaiactwebsite}. The Blueprint for an AI Bill of Rights outlines principles for the ethical development and deployment of AI technologies. It stresses the need for safe and effective AI systems, protection against algorithmic discrimination, data privacy, transparency, and the availability of human alternatives and recourse. Finally, the AI Risk Management Framework provides a guide to understanding the risks present in AI systems as well as a framework for designers, developers, and deployers of AI to evaluate and mitigate the risks relevant to the specific contexts of their system. 
These frameworks share a commitment to protecting individual rights, ensuring transparency, and promoting fairness in the use of technology, reflecting a global effort to ensure the responsible and trustworthy use of AI in real-world applications.

While these frameworks introduce and emphasize the necessity of comprehensive guidelines, there is a lack of clarity on how these guidelines can be effectively followed. Although extensive research exists on operationalizing various notions of trustworthiness, including transparency, fairness, and privacy, this information is often buried in dense technical papers that practitioners find difficult to navigate. For example, concepts like group fairness—which ensures that different demographic groups are treated equally—and individual fairness—which ensures that similar individuals receive similar treatment—are crucial but complex. Similarly, there are different approaches to enabling model transparency, such as building simpler models that are inherently interpretable versus constructing post hoc explanations for more complex models. The detailed methodologies for implementing these types of fairness and transparency are often hidden in academic literature, making practical enforcement challenging. Furthermore, the rapid pace of technological advancements complicates the consistent application and enforcement of regulatory frameworks, potentially undermining the very protections these frameworks aim to provide. 

In this write-up, we aim to address this shortcoming by providing an overview of various existing methods in literature  for operationalizing the aforementioned regulatory principles. For clarity and precision, we focus specifically on the Blueprint for an AI Bill of Rights in this document as it begins to lay out how these practices should look based on existing research. We present easy-to-understand summaries of state-of-the-art literature and highlight the gaps that exist between the regulatory guidelines and current AI research, including the trade-offs that emerge when implementing these principles. We hope that this work aids practitioners in understanding how to operationalize the regulatory guidelines outlined in the Blueprint for an AI Bill of Rights and also provides researchers and policy makers with a list of critical open problems and gaps between regulations and state-of-the-art AI research. Our goal is to strike a balance between providing deep technical understanding while concisely developing useful intuition about methods for operationalizing the AI Bill of Rights. We recognize that research in the fields surrounding these principles is evolving rapidly; however, we do not aim to provide an exhaustive survey of available methods but rather point to relevant starting points for a practitioner who wants to comply with regulation. As this is a working paper, we solicit feedback on various aspects of this document, including commentary on any gaps in existing policy, to further dialogue between practitioners, researchers, and policymakers.

This document is structured into multiple sections, each focusing on a specific guideline outlined in the Blueprint. For example, Section \ref{sec:safe_and_effective} focuses on Safe and Effective Systems, while Section \ref{sec:algorithmic_discrimination} addresses Algorithmic Discrimination Protection, and so on. Each section begins with a brief summary of the regulatory principle outlined in the Blueprint, followed by examples illustrating how the guideline applies to real-world applications. We then describe existing research that can aid in operationalizing the guideline and discuss specific considerations for generative AI. Each section concludes with an identification of gaps in the research of AI systems. At the end of the write-up, in section \ref{sec:intersections} hat examines the interplay between various regulatory guidelines and how to navigate the challenges that arise when operationalizing multiple guidelines simultaneously. 

\section{Safe and Effective Systems}
\label{sec:safe_and_effective}
\emph{Safe and Effective Systems} ensures that systems are designed to limit potential harms, are tested rigorously before deployment, and are monitored after deployment. This principle asks that developers of automated systems consult with the communities that will be affected by the system, as well as all other stakeholders, to confirm that all possible concerns and risks are identified before development. Then, during development, extensive testing is required to guarantee that the automated system is safe, with the expectation that system deployment may be delayed or prevented if tests are not passed. In accordance with this principle, developers should confirm that excessive data is not collected and that all data is used appropriately and safely.  Finally, ensuring the long-term safety and efficacy of an automated system requires continuous monitoring and reporting after deployment,  and mechanisms for adaptation and updates if necessary.

Regardless of the intentions of the designers of any technology, users can always intentionally or accidentally abuse technological systems. Such potential risks must be identified and considered during the design and development of automated systems to limit unintended harms and prevent ill use after deployment. The recent increased pace of automated system adoption has exacerbated the opportunity for misuse, as decreased time has been spent studying and testing systems in controlled environments before deployment. For example, while automated plagiarism detection systems are built to uphold academic integrity and reduce the workload of teaching staff, the effect that their immediate widespread adoption has had on students’ conceptualizations of collaboration and originality is still unknown. Furthermore, such software can also be actively abused to further personal agendas, such as targeting and discrediting specific academics by quickly searching large bodies of work \citep{purtill2024}. As another example, consider the exponential growth of utilization of large language models by the public. The adoption of these systems outpaced the recognition and dissemination of their limitations, and many users placed too much trust in the factuality and abilities of these models, resulting in hallucinated outputs being treated as facts and used in legal proceedings and academic work \citep{dahl2024large, bowman2022, buchanan2024chatgpt}. 

Developing safe and effective systems begins with a responsible design process that consults all stakeholders and identifies potential social, environmental, and individual risks a system might pose. To engage in responsible design, developers need to determine who to involve in the design process and what is at stake for those affected by a system. Then, depending on the context and impact of the automated system, practitioners can select from a wide body of literature on participatory approaches to design \citep{delgado2021stakeholder, zytko2022participatory}. Early work in user-centered design considers the needs of users during the design process \citep{norman1986user}, while service design considers a larger set of stakeholders impacted by a system \citep{forlizzi2018moving}. Participatory design practices dive deeper, understanding inherent power balances between providers and users in the design process and seeks to navigate and limit the impact of these political structures \citep{muller1993participatory, simonsen2013routledge}. Action research involves stakeholders as co-researchers and developers of systems \citep{elliot1991action}, while value-sensitive design emphasizes the ethical values of stakeholders during the design process. Finally, mechanism design proposes methods rooted in social choice theory \citep{arrow2012social} to quantitatively aggregate stakeholder preferences \citep{finocchiaro2021bridging}. Methodologies and tools from these various design approaches can guide reflective and critical thinking, help to identify unintended consequences of systems, provide means to address such problems, and amplify underrepresented communities and voices in the design process. 

After determining an appropriate design process, practitioners must consider what data, if any, is necessary to construct their automated system. If data is required to train or design a system, then this data must be of high quality and an accurate representation of the population that will eventually be affected by the system after deployment. For instance, a movie recommendation system trained on adult user preferences may not correctly learn the preferences of children and may recommend inappropriate content, so the developers of this system should consider collecting data on child movie preferences. As mentioned previously, developing a deep understanding of stakeholders and impacted communities can help indicate what types of data to collect and from whom to ensure relevance. In addition to curating high-quality data, deployers of automated systems should be aware of the risks of data reuse---especially in high stakes settings---where data is collected for one purpose and then used in a completely different context, especially by another group. For instance, genetic information collected by the FBI for security purposes could be used by researchers to train a genomics model, despite this data being collected in the context of criminal investigations and thus likely containing demographic biases that could drastically impact performance in the reused setting. If a biased dataset becomes a standard benchmark in academic research or industry, it will result in a large scale shift to systems reflecting and potentially perpetuating these biases when deployed. Thus, in high-stakes settings, this data should come with explicit reports on the context within which data was collected and its purpose.

For a system to be safe, it must be robust to a wide variety of inputs and perform in a predictable manner. Literature in adversarial robustness and distribution shifts addresses this issue in two different settings. Adversarial robustness considers a malicious adversary who is trying to break a system’s behavior with small, unnoticeable perturbations of inputs. Adversarial training, or training a machine learning model on a mixture of ground-truth and adversarial examples, provides a potential solution, minimizing the impact of malicious attacks \citep{goodfellow2014explaining}. Furthermore, several methods have been proposed that provide certified robustness: by training with these methods, a system can be guaranteed to be robust up to a certain level of attack \citep{pmlr-v97-cohen19c, hein2017formal}. Distribution shift research considers the case where a machine learning model receives unexpected inputs, specifically the case where the input data to a model after deployment (i.e. the testing or inference data) is different from the data it was trained on. The change in performance due to distribution shift can be counteracted using techniques from Distributionally Robust Optimization (DRO) \citep{Rahimian_2022}. Depending on the application, a system may be at risk from adversarial attacks, distribution shifts, or both. Developers and designers should evaluate these risks and accordingly consider methods for improving their systems' robustness. 

Both before and after deployment, automated systems must be tested and continually monitored to ensure they remain safe and effective. Testing should occur internally, by the developers of an automated system, and externally, through public audits and independent reporting. Internal tests should consider edge cases and measure potential consequences and harmful outputs of the system. They might also include user studies of the system to evaluate limitations, effects on various stakeholders, and compliance with the principles of explainability, protection against discrimination, and privacy. An example framework for internal testing and reporting is SMACTR, which provides an end-to-end auditing approach through all stages of development \citep{raji2020}. After deployment, external tests and auditing should be performed to verify the results of internal testing, to generate public reports of system behavior, and to consider applications beyond what the designers, developers, and deployers may have tested a system for. These tests should be standardized so consumers can easily understand and compare various offered automated services and make informed decisions about how to use these systems. Furthermore, in the case that an automated system violates external testing, the provider of the said system should flag this violation to all users and take appropriate actions to remedy the system, including the possibility of removing the system from deployment.

\paragraph{Considerations for Generative AI.}
While many of the concerns for predictive systems remain for generative systems, designers, developers, and deployers must take extra precautionary steps to ensure generative systems are safe and effective. In particular, given the wide variety of tasks and applications that a single generative model can be adapted and used for such as writing code, answering questions, and generating images, there are significantly more ways that generative systems can be misused by users to cause harm. It is imperative that designers and developers consider all of these potential downstream use cases before deployment. For example, while image generation models can be used to create art, edit images, and recreate photographs, they have also been leveraged to create deepfakes and explicit content. Language models can be used to spread misinformation by generating highly plausible incorrect information or by impersonating humans on social media. 
Furthermore, generative models are known to frequently produce ``hallucinations'' \citep{zhang2023siren}, or factually incorrect outputs, which pose additional risks in high-stakes settings. While retrieval augmented generation (RAG) can increase the factuality of models, this problem is still ongoing \cite{khandelwal2019generalization, lewis2020retrieval}.
The creators of generative models must provide detailed testing and reporting on the potential harms a model could cause to users under normal circumstances as well as risks to society when used by an adversary. This could includes listing out known instances of factual incorrectness, as well as the year of knowledge to which a model is trained. Current solutions to ensure the safe and effective behavior of generative models involve fine-tuning them to align with human preferences, also known as Reinforcement Learning from Human Feedback (RLHF) \citep{ouyang2022training}, in addition to traditional fine-tuning strategies. Other training-free interventions include developing prompting strategies such as system prompts, which instruct an LLM with chat capabilities how to interact with a user, and model editing methods which try to intervene on model weights and representations to improve factual correctness \citep{hase2021language}.

\paragraph{Considerations for Reporting.} To report that an automated system is safe and effective, practitioners should consider the following. First, designers should provide a set of potentially impacted populations, a justification for their consideration in the design process, and how this group contributed to a participatory design framework, if appropriate. In addition, the potential risks and harms of a system should be publicized, using assessment tools such as the AI Risk Management Framework \citep{ai2023artificial}. Next, to provide transparency on data quality, developers should use approaches such as the datasheets framework \citep{gebru2021datasheets}, which asks questions about the motivation for collecting a dataset, its composition, how it was collected, how it was preprocessed, how it should be used, and how it should be distributed and maintained. To verify a system is robust, a report should note if the system was trained with any robust optimization practices such as DRO or adversarial training, as well as if it has any level of certified robustness.  Finally, system performance on testing before and during deployment should be made publicly available, as well as information regarding the testing itself.

\paragraph{Gaps in Research.} Research on methods for ensuring safe and effective systems in practice is still ongoing, and there exist many areas for future work. Collecting high-quality data often requires manual selection, labeling, and cleaning, but as automated systems become larger and require more training data, methods to automate this process remain an important open problem. Furthermore, the tendency of generative models to ``hallucinate'' by presenting false information as facts remains a significant ongoing challenge. Finally, detecting generated content, such as deepfakes and false propaganda, also constitute a gap in the literature, whose resolution could prevent misuse of models for malicious purposes. 

\section{Algorithmic Discrimination Protection}
\label{sec:algorithmic_discrimination}
\emph{Algorithmic Discrimination Protection} states that users should be protected from unjustified differential or harmful treatment by an automated system on the basis of race, color, ethnicity, sex, religion, age, national origin, disability, veteran status, genetic information, or any other identity protected in existing legislation. Algorithmic discrimination protections should be considered and implemented during the design, development, and deployment of automated systems to mitigate the disparate treatment of individuals and groups. These protections include ensuring accurate representation (in design input and data collection) of the peoples impacted by an algorithm, active interventions to develop a system that minimizes discrimination, and continual testing and monitoring to limit discrimination after system deployment. Furthermore, the principle emphasizes the need for auditing and reporting of algorithms for disparate impacts and treatment.

As automated decision-making systems are increasingly integrated into all aspects of life, including many high-stakes domains such as healthcare, hiring, and credit, preventing algorithmic discrimination is pivotal in protecting the civil liberties and rights of the American people. Current protections against discrimination must be expanded to the digital sphere to ensure that decisions made not just by people but also by automated systems result in fair treatment and protection of all. This is especially true in settings where algorithmic decisions can have high or long-term impact on the quality of life of individuals. For instance, a bank using an automated loan approval system determines which members of the community are able to start a small business or afford college. These opportunities have lasting impacts on wealth-building and the financial prosperity of families and communities. While some assume that automating the loan approval process removes human subjectivity and bias, many systems–especially machine learning systems–learn the biases of the human data they are trained on, and thus can perpetuate or even exacerbate those biases. 

While the principle of algorithmic discrimination protections highlights the importance of preventing unjust treatment by automated systems, implementing this principle may be very difficult in practice. Discrimination protections begin at the design stage, where the system designers must ask whether the fundamental purpose, inputs, and potential outcomes of the system are perceived to be fair. In many cases, the decision of whether or not to give sensitive attributes as inputs to the system is nontrivial. Ignoring sensitive attributes can avoid discrimination, but in some cases,  considering and protecting these attributes may be more fair. For example, an automated grading system likely should not consider the demographic information associated with the authors of its inputs, and as such would be more fair if it was unaware. On the other hand, in the case of a system that awards need-based scholarships, the system might be more fair overall if it considers applicants and their demographics holistically. 

The second step to building a fair system is to pick the definition of fairness that best suits the use case, outcomes, and stakeholders of the system. These definitions can be broadly categorized into group fairness and individual fairness. Group fairness definitions evaluate the differences in system performance across different demographic groups. Popular examples of this include demographic parity \citep{dworkfairness2012}, which checks for the difference between selection rates of various groups, equalized odds \citep{hardt_equality}, which evaluates the difference between both true positive and true negative outcomes for various groups, equality of opportunity \citep{hardt_equality}, which considers the difference between only true positive outcomes for groups, and multi-calibration/multi-accuracy, which considers performance on arbitrary partitions of subgroups \citep{hebert-johnson18a, kim2019multiaccuracy}. Causal fairness metrics use methods from causal inference to study the effect that sensitive group attributes and demographic labels have on the outcome or prediction of an automated system. Popular metrics include path-specific fairness \citep{chiappa2019path}, which considers fair paths in a causal graph,  and counterfactual fairness \citep{kusner2017counterfactual}, which considers how the outcome changes if the sensitive attribute is changed. On the other hand, individual fairness formalizes the idea that similar individuals, or similar inputs to a system, should be treated similarly, meaning they should result in similar outputs by the system. To compare these measures, consider a company that builds an automated system to hire candidates for four open managerial positions. The applicant pool considered by the system is uniformly distributed among four age groups. As such, demographic parity is satisfied when an applicant from each age group is hired. However, if half of the competitive or qualified applicants come from just one age group, then equality of opportunity would only be satisfied if half of the positions are filled with applicants from that age group. Counterfactual fairness, on the other hand, is met if the system’s recommendation does not change even if the age of an applicant is arbitrarily altered. 

After selecting a definition of fairness, the developers must then determine if their system is fair under this definition, and if not, how to intervene to protect against discrimination. As such, the last two steps for building a fair system are fairness evaluation and intervention. First, designers should evaluate their automated system by surveying the outcomes on past data to determine if they satisfy the definition chosen. Here, it is important to select data from a representative stakeholder population to gain a better understanding of the bias in the system. After evaluation, there are many fairness interventions a practitioner can select from, but they can be broadly categorized into pre-processing, in-processing, and post-processing methods. Pre-processing considers steps one can take before the construction or training of an automated system, such as balancing a dataset \citep{calmon2017optimized} or sampling more data from an underrepresented population \citep{pmlr-v139-rolf21a}. In-processing methods consider interventions applied during the training of an algorithm, such as regularization in the optimization of machine learning models \citep{zafar2017fairness}. Finally, post-processing methods propose modifications one can apply post-hoc to modify an arbitrary system into a fair one by correcting for bias over various demographic groups \citep{agarwal2018reductions}. As society continues to evolve over time, population demographics can shift and what we define as a ''sensitive attribute" can also change. Thus, continuous monitoring of fairness and reevaluation of what attributes should be protected is imperative in any automated system.

\paragraph{Considerations for Generative AI.}
The majority of formulations of fairness in existing literature focus on predictive systems, and thus are measured in terms of outcomes for individuals or subgroups. As such, these methods may not translate directly to generative systems. However, bias is still a pressing issue in generative AI, particularly as systems and models scale in size, as outputs may perpetuate stereotypes or may be skewed towards certain populations. For example, language models tend to associate certain professions with a specific gender, and image generation models may tend to produce people of a certain age or demographic more than others \citep{luccioni_stablebias}. Another consideration in generative settings is whether models should explicitly consider sensitive attributes (such as gender) and balance generations along these dimensions, or if models should be debiased to ignore sensitive attributes during generation.
Existing solutions include fine-tuning models with various fairness objectives given labelled datasets or specifying desired behavior, such as via prompting, to ensure unbiased outputs. Reinforcement learning from human feedback (RLHF) is a widely-used example of such finetuning, aimed at increasing alignment between humans and models across multiple axes, including fairness \cite{ouyang2022training}. 
However, these methods provide no guarantees that the model will no longer have biases, and recent work has shown that debiasing and alignment through fine-tuning and prompting can easily be undone by users \citep{zhan2023removing}. The demonstrated potential for discrimination by generative models motivate the need for new metrics of fairness and bias in generated outputs as well as novel methods to ensure that models generate appropriately diverse outputs free from stereotypes. 

\paragraph{Considerations for Reporting.} Reporting the performance of a system across various populations is important for enabling stakeholders to make informed decisions regarding their interactions with automated systems, especially when they may be discriminative. When collecting a dataset for public usage or for the construction of a downstream system, developers should conduct an assessment of the distribution of different demographics in the dataset. Reporting demographic information will allow developers to select the most relevant dataset when creating a system and will allow users to understand if systems trained on this dataset will treat them fairly. In addition, providers should report what potentially sensitive information is being used in a decision making system, as well as aggregate performance on applicable fairness metrics. Finally, if specific methods are used to ensure a system is fair, these methods and what sensitive groups they seek to protect should be named in the system report.

\paragraph{Gaps in Research.} Despite the robust set of fairness metrics and methods to protect fairness, there are still many gaps between system capabilities and policy goals. First, the tradeoff between fairness and accuracy draws a Pareto curve that practitioners can navigate, but selecting a specific point on this curve can be a difficult and arbitrary process, posing political and ethical questions. Furthermore, simply optimizing for a fairness metric does not necessarily result in a system that is fair. It is important to consult all potential stakeholders, especially those directly affected by an automated system, to understand what “fairness” means to them and how it can be achieved with the said system. Finally, recent work has focused on the arbitrariness of machine learning systems, where there exist multiple possible models with different individual outcomes but the same overall accuracy. If two models have equivalent group- or population-level performance, but one gives a specific individual a loan and the other does not, how should a practitioner or policymaker determine which model is more fair? This is also known as the predictive multiplicity problem, and resolving it is still an open question in algorithmic fairness literature. 

\section{Data Privacy}
\label{sec:data_privacy}
Data privacy states that only minimal and strictly necessary data should be collected by designers, developers, and deployers of automated systems, and that permission should be asked for and respected regarding the collection, use, access, transfer, and deletion of personal data. These protections should be provided by default and implemented in good faith, without unnecessary and obfuscatory interfaces or language that burden or confuse users. Furthermore, in sensitive domains such as health, work, education, finance, and criminal justice, additional protections and restrictions should be placed such that only the most necessary data is used and is protected by ethical review and use prohibitions. Surveillance technology should be overseen to scope impact and limit potential harm and should be completely avoided in high-stakes settings where it is likely to limit rights and opportunities. Finally, the collection and use of private data should always come with reporting to confirm that data privacy is being protected and its impact is well understood.

Privacy has always been a core component of American civil liberties, but its interpretation in the digital age is a new and evolving subject. With the extreme popularity of big data, increased surveillance and data collection can lead to a variety of infringements on user privacy. For example, social media companies may collect user data to sell to third-party agencies, such as a hiring agency, who may then try to infer their candidates’ political leanings. While in this case the hiring agency is intentionally trying to recover private data, many automated systems can still inadvertently leak such data if explicit privacy protections are not put in place. For instance, many websites leverage browsing behavior for targeted advertisements, often without the knowledge of their users. The subsequent advertisements can inadvertently reveal protected and private identities, such as gender and sexual orientation or pregnancy status \citep{target2012}. Even when users have consented to data collection, they should still have agency over the use of the data, they should be informed about potential risks, and data collection should still be minimized in scope to prevent harm, inadvertent or adversarial. Thus, it is important that developers understand the data privacy impacts of their automated systems and enact limitations and safeguards against overcollection and misuse.

While privacy is a highly valued and agreed upon right in the United States, both defining privacy and ensuring it are very difficult in modern society, with the digital world being no different. First, user data must be collected and stored in a nonintrusive and safe manner. During system design, the data privacy principle states that consent should be clearly asked for and given before a system collects or uses your data. This requires that the method of asking for permission is easy to understand and use, prioritizes privacy by default, and clearly explains how data will be used after collection. For example, a website that leverages user cookies should ask for consent from the site user before collecting the cookies, and should not by default suggest that the user share unnecessary cookies for the website to sell to advertising agents. If an organization collects personal data, it must also be stored in a secure manner to protect user privacy. In distributed settings, methods such as federated learning \citep{federatedlearning} can be used to limit the transfer and storage of personal data. This distributed training approach only communicates model updates to central servers, rather than the training data itself, allowing for personal data to remain private and secure. 

In addition to collecting the minimal data necessary to maximize user privacy, it is important to ensure data privacy is maintained on deployed systems with public access. For all data-based systems, careful thought should be put into how systems may inadvertently reveal their data to other users or adversaries. In the case of machine learning systems, if the weights of a model are made public, then training data can often be recovered, violating the privacy of those who provided the data \citep{nasr2019comprehensive}. This is true even if adversaries do not have access to the weights and instead are able to query the model many times with a variety of inputs, also known as a Membership Inference Attack (MIA) \citep{shokri2017}. As such, in high-risk domains, developers should consider additional security measures when training machine learning models, such as methods in differential privacy. Differential Privacy \citep{dwork2006calibrating} uses statistical guarantees to ensure that an adversary querying a model is unable to distinguish between data points that were or were not present in the model's training data. This is generally achieved by adding noise to the objective function \citep{chaudhuri2011differentially} during learning or to system outputs during inference \citep{dwork2006calibrating}. In deep networks, a popular method is Differentially-Private Stochastic Gradient Descent (DP-SGD) which adds noise and clips gradient values during backpropagation to achieve privacy \citep{abadi2016deep}.

After implementing privacy protections during the design of an automated system, developers should provide mechanisms to give users agency over the data they provide. The data privacy principle states that users should be able to access all data about themselves as well as other metadata, such as who else has access to this data. Along with access to this information, developers should also provide pathways for users to request this data be withdrawn from use or corrected. Given that machine learning models can leak training data, developers need to ensure data can be deleted from both storage bases and from the models themselves.  To do so, developers can retrain their models, but retraining for every user modification or deletion can be computationally inefficient or infeasible. This has motivated a body of research on machine unlearning, which seeks to find efficient approximations to retraining using methods from security and privacy \citep{bourtoule2021machine, guo2019certified}.

\paragraph{Considerations for Generative AI.}
Generative AI poses significant threats to data privacy beyond those of predictive models because systems can replicate training data during generation. This phenomenon is frequently referred to as “memorization” and has been shown to be a problem across many modalities that scales with model size. For instance, with targeted prompting, researchers have been able to extract personal information found in the training data of models such as addresses and phone numbers \citep{carlini2021extracting}. These generative models require large-scale training datasets, often scraped from the internet, and even if this data is publicly accessible, it may still be considered private or sensitive, such as a leak of personal passwords or emails. 
Existing solutions to this problem include applying traditional privacy methods, such as DP-SGD, to generative models during fine-tuning or using unlearning methods to remove sensitive information after the model has been trained. Unlearning methods seek to make models forget individual training data and can be used to remove sensitive samples, while fine tuning can be used to take a model pretrained on a public dataset and improve it with private data in a safe manner. However, the large number of parameters in generative models make privacy methods generally inefficient and costly in terms of performance. Recent works have also explored how to detect memorization through data attribution methods or influence functions, which determine the impact of training data on an output. However, these attribution methods are expensive in practice especially at the trillion-token scale of language model training datasets. 

\paragraph{Considerations for Reporting.} When communicating the privacy of an automated system, there are a few important metrics to consider. Reporting the kind of data, such as medical information, income, or demographic attributes, a system takes as input or stores is necessary for users and regulators to understand the risks of a security breach. In addition, if any specific methods are used to preserve privacy, such as federated learning or differential privacy, these methods should be reported, and in the case of differential privacy, its privacy leakage (a quantification of how private the resulting system is) should also be made public. Furthermore, if a system advertises the ability to unlearn users, the specific method and whether it achieves exact unlearning or approximate should be noted. 

\paragraph{Gaps in Research.} While the field of privacy research in digital systems is an active and large area of research, there are still gaps between our policy goals and technological capabilities. First, while there are many mechanisms for achieving differential privacy, these often come at a cost to accuracy that scales with model size as well as the number of training steps. Furthermore, machine unlearning methods only provide theoretical guarantees for a select few model architectures, and thus often do not apply to many popular large deep neural networks. Finally, as generative AI becomes increasingly prevalent and as models have scaled up in size and use, the potential for models to inadvertently reveal training data during inference has also increased. Understanding of when and why models output memorized samples is still unclear, making it difficult to prevent this behavior, despite implications for privacy and copyright law. 

\section{Notice and Explanation}
\label{sec:notice_and_explanation}
Notice and explanation states that users who are at the receiving end of outcomes from automated systems should be clearly informed that an automated system is being used, and they should also be provided with an explanation of how and why the system contributed to the outcome. Furthermore, this principle asserts that designers, developers, and deployers of automated systems must provide documentation about system behavior, the role of automation, and information regarding the individuals and organizations responsible for the system. Most importantly, stakeholders are entitled to clear, timely, accessible, and valid explanations of the outcomes and decisions made by automated systems.

The principle of notice and explanation provides critical guidelines that are essential in today’s world where automated systems are powering diverse real-world applications involving high-stakes decisions. For instance, automated systems are being employed in hiring, credit, and courtrooms in ways that profoundly impact the lives of the American public. However, this impact may not always be obvious given the minimal transparency around such systems. For example, a job applicant might not know whether a recruiter rejected their resume or if instead a hiring algorithm placed them at the bottom of its list. Similarly, a defendant should know if a judge that denied their bail was given a recommendation by an automated system. However, even knowing that a system was used in a decision-making process is not enough, and affected individuals must also be provided with reasons or explanations for their outcomes. The lack of transparency around automated systems prevents individuals who are at the receiving end of outcomes from contesting or appealing negative decisions impacting their lives. The principle of notice and explanation outlines procedural guidelines to address the aforementioned challenges and promote transparency in the adoption and use of automated systems employed in real-world applications. 

While this principle emphasizes the need for explanations, it does not provide prescriptive guidance in terms of what approaches or methodologies can be used to operationalize these guidelines. For example, providing explanations for outcomes produced by automated systems typically involves understanding the rationale behind the predictions of the underlying machine learning models. This is a technically challenging problem that has been studied extensively in machine learning and statistics literature \citep{gilpin2018explaining} under the broad umbrella of interpretable or explainable machine learning. 
Prior research on this topic has proposed two broad solutions to the problem of understanding the behavior or predictions of ML models. The first approach involves developing models that are inherently interpretable and are therefore easy to understand by design. For example, simple predictive models such as linear or logistic regression, which output feature coefficients that can be interpreted as importance scores assigned to individual features, and (shallow) decision trees/lists/sets, which output rules that can be used to make predictions on data instances are often considered inherently interpretable models \citep{lakkaraju2016interpretable, rudin2022interpretable}. Such models are easily understandable to diverse classes of end users and stakeholders with little to no expertise in machine learning and statistics. 
While the aforementioned simple models are readily interpretable, prior works have demonstrated that these models may not be as accurate as their complex counterparts that are hard to interpret (e.g., deep neural networks or ensemble models such as random forests and boosted trees). To this end, a second approach to understanding the behavior of ML models, frequently referred to as ``post hoc" explainability, has risen in popularity. These methods allow for the use of complex or black-box models that are hard to interpret but accurate in real-world applications by constructing “post hoc” explanations of these models during inference. Post hoc explanations are typically constructed by approximating the local (around the vicinity of a single data point) \citep{han2022explanation} or global (complete model) behavior of complex models using simpler and easy to interpret constructs. While post hoc explanations may take several forms, some of the most common post hoc explanations found in literature are feature importance based explanations \citep{ribeiro2016should, lundberg2017unified, sundararajan2017axiomatic, selvaraju2016grad, smilkov2017smoothgrad} and counterfactual explanations \citep{wachter2017counterfactual}. Feature importance based explanations assign importance scores to input features (e.g. highlighting important pixels in an image), whereas counterfactual explanations describe what features need to be changed and by how much in order to change the model's prediction (e.g. earning an additional \$$10,000$ of income will qualify an applicant for a loan that they were previously denied). As such, feature importance based explanations are used to provide informative explanations by revealing the features that drive an outcome while counterfactual explanations are more prescriptive and are geared towards allowing for the possibility of recourse. 

While inherently interpretable models are not as accurate as their complex counterparts, they are readily understandable to a variety of stakeholders and their explanations are generally considered valid and faithful. In contrast, post hoc explanations provide approximations of the behavior of underlying models and therefore may not always be accurate  \citep{rudin2019stop, han2022explanation}. Furthermore, these explanation methods may misrepresent model behavior and can be fooled with adversarial techniques \citep{adebayo2018sanity, dombrowski2019explanations, slack2020fooling}. Given this, it becomes important to rigorously evaluate the validity and quality of post hoc explanations before relying on them or prescribing them in real-world applications. The correctness of a counterfactual explanation is evaluated by ascertaining that making the changes prescribed by the explanation actually results in a positive outcome from the model. Evaluating the validity of feature importance based explanations is more challenging in practice due to the lack of authoritative ground truth i.e., we do not necessarily know which features are driving a specific model prediction \citep{bhalla2023verifiable}. To this end, several proxy metrics/heuristics that assess the sensitivity of model output to changes in input features have been proposed, such as perturbing the features deemed as important by an explanation and analyzing how the model’s output changes in response \citep{fong2017interpretable}. 

\paragraph{Considerations for Generative AI.}
Given the complexity of the output space of generative models, such as high-dimensional images or large text outputs, many existing methods for producing explanations in predictive models may not be directly transferable to large generative models. This has led to a variety of new methods for interpreting generative outputs. Data attribution methods seek to explain generated outputs with respect to the training data points that influenced them \citep{grosse2023studying, ilyas2022datamodels}. Methods in mechanistic interpretability look into the internals of models and explain outputs through activations of certain neurons, circuits, and layers that have been attributed to certain behaviors \citep{wang2022interpretability, templeton2024scaling}. Concept-based explanations seek to generate semantic explanations of the concepts present in a task, representation, or generation \citep{bhalla2024interpreting}. Finally, language models have been use to explain themselves, through methods such as chain-of-thought prompting, and other models through constructed linear maps. 
Chain-of-thought prompting \citep{wei2022chain} asks models to provide step-by-step explanations of their reasoning, both improving performance and providing built-in explanations with a generation. However, it is unclear whether generated natural language explanations are faithful to the internal behavior of a model \citep{turpin2024language} or just learned responses from internet training data or RLHF fine-tuning. 

\paragraph{Considerations for Reporting.} While explanations themselves may be considered reports of how a system functions, it is also important for designers, developers, and deployers to report how these explanations are generated. For instance, the exact explanation generation method and the validity of the resulting explanations should be reported to stakeholders. If prescriptive explanations are provided, the sensitivity of an explanation should also be reported (e.g. an applicant should increase their income by \$10,000-15,000, not more or less, to be approved for a loan). Finally, notices and explanations must be kept up-to-date, and any changes made to the system should be reported to stakeholders and users, particularly if it changes their outcome or the explanation provided to them.  

\paragraph{Gaps in Research.} Despite the extensive literature exploring various forms of explainability and interpretability in multiple modalities, domains, and applications, there still exist some gaps in research. First, many of the explanations yielded by these methods may not be perfectly faithful, meaning they may provide an approximation of the exact reason why an automated system produced a certain output. Furthermore, these methods do not frequently report the degree of this unfaithfulness, or uncertainty, which would allow users of the system to assess whether or not an explanation is trustworthy. Another open area of concern is how to report explanations in a manner that is meaningful, useful, and interpretable to users of various degrees of familiarity and expertise with the system and its domain, while still remaining faithful and informative. For example, if an explanation highlights all the pixels in an X-ray image as being relevant to a prediction, is the explanation still actionable or meaningful to a user, even if it is faithful? Recent work on evaluating the usefulness of various explanation forms in real-world applications has revealed that explanations can frequently be misleading, mistrusted, unhelpful, and sometimes even actively harmful \citep{lage2019human, hong2020human}. As such, it is imperative that both explanations and the manner in which they are reported to users are given careful consideration and are properly evaluated for both empirical validity and effectiveness via user studies before the systems are fully deployed. 

\section{Human Alternatives, Consideration, and Fallback}
\label{sec:human_alternatives}
\textit{Human Alternatives, Consideration, and Fallback} proposes that, when applicable, users should be able to refuse use of an automated system and request a human alternative as a replacement. Importantly, the human alternatives must be easily accessible; users should know whether or not they are interacting with an automated system, in what capacity, and what the purpose and impact of the system are, so that they can decide whether they need to request a human alternative. Furthermore, the process for requesting a human alternative should be straightforward. Human fallbacks should be timely and should also be available after the use of an automated system, such that they can address any problems that arise or remedy faulty decisions made by the automated system. The people behind these alternatives should be effective and properly trained for the task the automated system performs. Providing human fallback is most important in sensitive domains such as medicine, finance, and criminal justice, where automated decisions can have drastic impacts on their stakeholders, and where human consideration and empathy are of utmost concern. 

There are many settings in which users may prefer or need a human alternative to an automated system. Consider the scenario of a medical clinic using a machine learning model to screen for risk of heart disease. If this model has a known bias towards a specific subgroup, patients of that population may wish to have a human medical professional consider their case instead of the automated system. In this situation, they should be provided with an alternative if they request one, or they should be able to have a human medical professional check the output of the model and verify that the explanation of the decision made by the automated system is correct. Note that there also exist settings in which human alternatives may not be necessary or appropriate, such as in settings where automated systems have minimal direct impact on their users and do not pose any risks (e.g. online machine translators, movie recommendation systems). 

Implementing human alternatives is relatively straightforward: wherever a practitioner deploys an automated system, they can add an option for users to select a human operator instead. However, an important consideration in the implementation of a human fallback is how much users trust a system. The perceived “impartiality” of automatic systems and their perceived inability to make mistakes may result in users being inhibited from requesting human fallback even when it could be in their best interests. The context and behavior of an automated system can also alter its perceived trustworthiness. Research has shown that providing explanations, even if incorrect or misleading, can increase user trust in a system by a factor of almost 10 \citep{lakkaraju2020fool}. Furthermore, factors such as the speed of an automated system and its interface also impact a users likelihood to trust the system \citep{glikson2020human}. Users should be provided with sufficient information regarding the benefits and harms of both the automated system and human alternative, such as wait times, accuracy, consistency, and more. Furthermore, there should not be negative consequences or punishments to users for requesting an alternative, and both the system and human alternatives should be made easy to use. Finally, in cases where a system or model is highly uncertain or is accurately aware that its output is incorrect or potentially harmful, the system itself should propose deferring to a human alternative, essentially auditing itself. This method is sometimes referred to as selective classification (or classification with a reject option), and while it has been shown to decrease error rates, recent work has also found that it may increase disparities among different subgroups \citep{geifman2017selective, jones2020selective}. 

\paragraph{Considerations for Generative AI.} The principle of providing human alternatives and fallbacks for automated systems remains the same for both predictive and generative systems.

\paragraph{Considerations for Reporting.} When reporting on the human alternatives offered by an automated system, there are few considerations. Practitioners should report exactly when users can request fallbacks during their interaction with an automated system, and what components of the system a human operator is replacing. In addition, practitioners should detail the training and expertise of the human operators. 
 
\paragraph{Gaps in Research.} It is unclear how providing human alternatives impacts the way in which users engage with automated systems. While there is research on the ways humans and automated systems can improve decision making processes together, the setting where they are posed as alternatives is less studied. To provide human alternatives in a way which prioritizes user rights, it is important to understand trust in automated systems and the effects of these interactions in terms of outcomes and other metrics.

\section{Intersections}
\label{sec:intersections}
While all of the goals enumerated in the Blueprint for an AI Bill of Rights are important independently, it is also necessary to consider the ways in which they can coexist or come into conflict with each other. For instance, optimizing an automated system to achieve auxiliary goals such as fairness or privacy generally comes at a cost to the performance of the system in terms of accuracy. In this section, we list potential conflicts that may arise in operationalizing the AI Bill of Rights to bring awareness to nuances operationalization.

While at first glance the principles of protecting against algorithmic discrimination and protecting data privacy may seem aligned, as they both prevent misuse of personal data, in practice these principles can come into conflict when developing an automated system. For example, optimizing for group fairness requires the collection of subgroup labels and sensitive attribute information, which may be at odds with the principle of privacy. Furthermore, the principle of data privacy states that users should be able to remove their data from automated systems, but if these deletions all come from one subgroup, this may result in models becoming unfair \citep{oesterling2023fair}. 

In many cases, increased transparency and explainability of automated systems can aid users in preventing algorithmic discrimination, particularly if explanations highlight biases of a model. However, there exist some challenges to ensuring systems are simultaneously fair and explainable. In general, explainability asks for predictors to be as simple or interpretable as possible, but oftentimes making a model more fair requires increasing its complexity \citep{kleinberg2019simplicity}. In some cases, forcing a model to be more interpretable by requiring simpler decision rules may bias it toward increased reliance on sensitive attributes.

Furthermore, protecting the right to explanation and protecting data privacy can often work against each other in practice. As explanations provide more information about the decision of a model, malicious actors are better able to reconstruct the model and can recover more information about training data if proper protections are not taken. Additionally, the right to be forgotten, allowing users to remove their data from automated systems, may result in explanation methods that fail to be faithful or accurate after a user’s data has been deleted from the system, invalidating past explanations and algorithmic recourse.
Implementing human fallbacks also requires consideration of its interactions with the other principles in the Blueprint for an AI Bill of Rights. For instance, providing explanations often inflates a user’s trust in a system, making them more likely to accept an automated decision instead of requesting human fallback \cite{papenmeier2019model}. However, proper reporting can allow consumers to better understand the risks involved with using an automated system and make a more educated choice about requesting a human alternative. Understanding if a system is fair, private, or robust to a wide variety of inputs can improve a user’s confidence in engaging with that system.

While there may exist tensions between the principles outlined in the Blueprint for an AI Bill of Rights, these principles also provide an avenue for navigating such tradeoffs. As outlined in the principle for Safe and Effective systems, by properly testing systems, practitioners can be aware of potential limitations and collisions between principles in practice. Furthermore, by reporting performance along various axes and being explicit about shortcomings, failure cases, and user protection tradeoffs of automated systems, users can make informed decisions about whether to use a system, disengage from it, or request human fallback.

\section{Reporting}
\label{sec:reporting}
In order to evaluate various systems on their compliance with the above principles, it is necessary to standardize a reporting structure for these systems. In addition, this standardization will allow consumers to fairly compare between automated systems to determine what to use, and whether to opt for human alternatives instead. A report should include information on the type of model used in an automated system, the data used to train it or the nature of this data (when possible), and the intention of the providers in creating this automated system. As part of these intentions, providers should highlight intended contexts for use, the expected types of users for the system, and the expected outcomes for these users. Extensive testing should also be conducted, using multiple sets of evaluation data. The results of this testing, in addition to a description of the evaluation data, should be provided to users. Results should include baseline performance metrics including loss and specific metrics such as fairness and privacy. In the above sections, we mention some considerations for reporting for each of the principles that should be included in a report.  Some example frameworks for reporting are the model cards framework \citep{mitchell2019model}, datasheets framework \citep{gebru2021datasheets} and the SMACTR auditing framework \citep{raji2020}. Finally, this reporting should be constantly-updated, with continual monitoring to track system performance over its whole lifecycle. Establishing a standardized reporting structure will require collaboration between academics and policymakers as well as regulation and enforcement to ensure adoption by system providers.

\section{Discussion}
In this work, we explore the intersection between the research, regulation, and practice of AI and address the tension between academic research and practitioner compliance with regulatory frameworks such as the AI Bill of Rights. To do so, we go section-by-section through the AI Bill of Rights and present a summary of each regulatory principle as well as an accessible set of state-of-the-art research that practitioners can use to implement regulatory principles. In addition, we highlight gaps between these principles and research to provide researchers with potential directions for future work. Finally, we consider how to document these practices for public access as well as potential trade-offs between various principles in the AI Bill of Rights. We welcome any and all feedback that helps further the purpose of this document to provide an accessible manual for practitioners while highlighting disconnects between research and regulation.

\bibliography{bibliography}

\end{document}